\pdfoutput=1
\documentclass[sigconf,nonacm]{acmart}

\usepackage[utf8]{inputenc}
\usepackage[T1]{fontenc}
\usepackage{microtype}
\usepackage[ngerman,english]{babel}
\usepackage[obeyFinal]{todonotes} 
\usepackage{listings} 
\usepackage{cleveref}
\usepackage{multicol}
\usepackage{glossaries}
\usepackage[commandnameprefix=always]{changes}
\usepackage{tikz}
\usepackage[binary-units,separate-uncertainty=true]{siunitx}
\usepackage{dblfloatfix}
\usepackage{subfig}
\usepackage[frozencache,cachedir=.]{minted}
\usepackage{import}
\usepackage{bm}

\usepackage{algorithm}
\usepackage{algpseudocode}

\makeatletter
\newbox\sf@box
\newenvironment{SubFloat}[2][]%
{\def\sf@one{#1}%
\def\sf@two{#2}%
\setbox\sf@box\hbox
\bgroup}%
{ \egroup
\ifx\@empty\sf@two\@empty\relax
\def\sf@two{\@empty}
\fi
\ifx\@empty\sf@one\@empty\relax
\subfloat[\sf@two]{\box\sf@box}%
\else
\subfloat[\sf@one][\sf@two]{\box\sf@box}%
\fi}
\makeatother

\definecolor{color0}{HTML}{1f77b4}  
\definecolor{color1}{HTML}{ff7f0e}  
\definecolor{color2}{HTML}{2ca02c}  
\definecolor{color3}{HTML}{d62728}  
\definecolor{color4}{HTML}{9467bd}  
\definecolor{color5}{HTML}{8c564b}  
\definecolor{color6}{HTML}{e377c2}  
\definecolor{color7}{HTML}{7f7f7f}  
\definecolor{color8}{HTML}{bcbd22}  
\definecolor{color9}{HTML}{17becf}  

\usetikzlibrary{calc,backgrounds,fit}

\newacronym{adc}{ADC}{analog-to-digital converter}
\newacronym{adex}{AdEx}{adaptive exponential integrate-and-fire}
\newacronym{afib}{AF}{atrial fibrillation}
\newacronym{ann}{ANN}{artificial neural network}
\newacronym{asic}{ASIC}{application-specific integrated circuit}
\newacronym{asicab}{\acrshort{asic} adapter \acrshort{pcb}}{\acrlong{asic} adapter \acrlong{pcb}}
\newacronym{api}{API}{application programming interface}
\newacronym{bmbf}{BMBF}{German Federal Ministry of Education and Research}
\newacronym{bptt}{BPTT}{backpropagation through time}
\newacronym{bss2}{\mbox{BSS-2}}{Brain\mbox{ScaleS-2}}
\newacronym{bss1}{\mbox{BSS-1}}{Brain\mbox{ScaleS-1}}
\newacronym{bss2os}{\gls{bss2} OS}{\gls{bss2} Operating System}
\newacronym{bss}{BSS}{BrainScaleS}
\newacronym{cdnn}{CDNN}{convolutional deep neural network}
\newacronym{cpu}{CPU}{central processing unit}
\newacronym{dfki}{DFKI}{German Research Centre for Artificial Intelligence}
\newacronym{dma}{DMA}{direct memory access}
\newacronym{dram}{DRAM}{dynamic random-access memory}
\newacronym{ecg}{ECG}{electrocardiogram}
\newacronym{fpga}{FPGA}{field-programmable gate array}
\newacronym{gbe}{GbE}{gigabit ethernet}
\newacronym{i2c}{I\textsuperscript{2}C}{Inter-Integrated Circuit}
\newacronym{ic}{IC}{integrated circuit}
\newacronym{isa}{ISA}{instruction set architecture}
\newacronym{itl}{ITL}{in-the-loop}
\newacronym{jit}{JIT}{just-in-time}
\newacronym{lvds}{LVDS}{low-voltage differential signaling}
\newacronym{lif}{LIF}{leaky-integrate and fire}
\newacronym{li}{LI}{leaky integrator}
\newacronym{mac}{MAC}{multiply–accumulate}
\newacronym{madc}{MADC}{membrane \acrshort{adc}}
\newacronym{mse}{MSE}{mean squared error}
\newacronym{cadc}{CADC}{columnar \acrshort{adc}}
\newacronym{pcb}{PCB}{printed circuit board}
\newacronym{ppu}{\acrshort{simd} \acrshort{cpu}}{\acrlong{simd} \acrlong{cpu}}
\newacronym{relu}{ReLU}{rectified linear unit}
\newacronym{rtl}{RTL}{Register Transfer Level}
\newacronym{gd}{GD}{gradient descent}
\newacronym{simd}{SIMD}{single instruction, multiple data}
\newacronym{snn}{SNN}{spiking neural network}
\newacronym{sodimm}{\mbox{SO-DIMM}}{small outline dual in-line memory module}
\newacronym{sram}{SRAM}{static random-access memory}
\newacronym{stdp}{STDP}{spike timing dependent plasticity}
\newacronym{stp}{STP}{short term plasticity}
\newacronym{rnn}{RNN}{recurrent neural network}
\newacronym{rsnn}{RSNN}{recurrent spiking neural network}
\newacronym{nasprop}{NASProp}{neuromorphic accumulative spike propagation}
\newacronym{vu}{VU}{vector unit}
\newacronym{udp}{UDP}{user datagram protocol}
\newacronym{cd}{CD}{continuous deployment}
\newacronym{ci}{CI}{continuous integration}
\newacronym{hpc}{HPC}{high-performance computing}
\newacronym{gpu}{GPU}{graphics processing unit}
\newacronym{usb}{USB}{universal serial bus}
\newacronym{sfnn}{SFNN}{spiking feed-forward neural network}
\newacronym{sfnnwlrf}{SFNNwLRF}{\gls{sfnn} with limited receptive field}
\newacronym{ttfs}{TTFS}{time-to-first spike}
\newacronym{scnn}{SCNN}{spiking convolutional neural network}
\newacronym{srnn}{SRNN}{spiking recurrent neural network}
\newacronym{dsnn}{DSNN}{deep spiking neural network}
\newacronym{dnn}{DNN}{deep neural network}
\newacronym{fpu}{FPU}{floating-point unit}
\newacronym{vjp}{VJP}{Vector Jacobian Product}
\newacronym{xla}{XLA}{Accelerated Linear Algebra}
\newacronym{mot}{MOT}{max over time}
\newacronym{nir}{NIR}{Neuromorphic Intermediate Representation}
\newacronym{fud}{F\&D}{Fast and Deep}

\begin{document}

\title[jaxsnn: Event-driven Gradient Estimation for Analog Neuromorphic Hardware]{jaxsnn: Event-driven Gradient Estimation for \\Analog Neuromorphic Hardware}

\author{Eric Müller}
\email{mueller@kip.uni-heidelberg.de}
\author{Moritz Althaus}%
\author{Elias Arnold}%
\author{Philipp Spilger}%
\affiliation{%
	\department{Kirchhoff-Institute for Physics}
	\institution{Heidelberg University}
	\country{Germany}%
}
\author{Christian Pehle}%
\affiliation{%
	\institution{Cold Spring Harbor Laboratory}
	\country{USA}
}
\author{Johannes Schemmel}
\affiliation{%
	\department{Kirchhoff-Institute for Physics}
	\institution{Heidelberg University}
	\country{Germany}
}

\renewcommand{\shortauthors}{E.\ Müller, M.\ Althaus, E.\ Arnold, P.\ Spilger, C.\ Pehle, and J.\ Schemmel}


\begin{abstract}
Traditional neuromorphic hardware architectures rely on event-driven computation, where the asynchronous transmission of events, such as spikes, triggers local computations within synapses and neurons.
While machine learning frameworks are commonly used for gradient-based training, their emphasis on dense data structures poses challenges for processing asynchronous data such as spike trains.
This problem is particularly pronounced for typical tensor data structures.
In this context, we present a novel library\footnote{\url{https://github.com/electronicvisions/jaxsnn}~\citep{jaxsnn2023github}} built on top of JAX, that departs from conventional machine learning frameworks by providing flexibility in the data structures used and the handling of time, while maintaining Autograd functionality and composability.
Our library facilitates the simulation of spiking neural networks and gradient estimation, with a focus on compatibility with time-continuous neuromorphic backends, such as the \acrlong{bss2} system, during the forward pass.
This approach opens avenues for more efficient and flexible training of spiking neural networks, bridging the gap between traditional neuromorphic architectures and contemporary machine learning frameworks.

\end{abstract}

\maketitle

\section{Introduction}\label{sec:introduction}

In recent years, the modeling of \glspl{snn} on neuromorphic systems has evolved, focusing on specialized interfaces for efficient model development~\citep{manna2023frameworks,mueller2022scalable_noeprint,mueller2022operating,rhodes2018spynnaker,ji2016neutrams,amir2013cognitive,dynapse2021nice_misc}.
In particular, the success of \glspl{dnn} has sparked the interest of modelers and created a desire for machine learning-friendly interfaces~\citep{manna2023frameworks}.
In this domain, many training approaches make use of software frameworks that are traditionally optimized for the handling of dense data.
However, neuromorphic systems typically operate on spikes~\citep{thakur2018mimicthebrain_nourl}:
events in time that occur asynchronously.
While building software support into typical machine learning libraries is well established for general-purpose processing units performing numerical operations~\citep{torchxla,lattner2021mlir,lohoff2023interfacing}, it remains an open research question for custom digital neuromorphic architectures~\citep{shrestha2022survey}.
This challenge is even more complex for analog neuromorphic systems with their continuous-time nature, making seamless integration a challenging task.

\begin{figure}[tb]
    \centering
	\begin{SubFloat}{\label{subfig:spike-encoding}}%
		\begin{minipage}[b]{.3\linewidth}
			\begin{minted}{python}
class Spike:
    neuron: int
    timestamp: float




			\end{minted}
		\end{minipage}
	\end{SubFloat}
	\hfill%
	\subfloat[\label{subfig:time-encoding}]{%
		\resizebox{.60\linewidth}{!}{%
			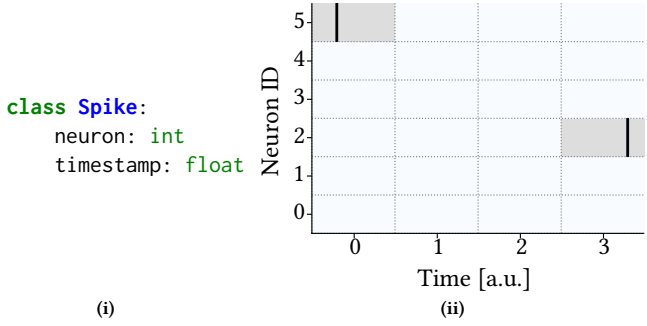%
		}%
	}%
    \caption{%
		\protect\subref{subfig:spike-encoding}: A spike comprises an identifier/label for the source neuron and a (high-resolution) time stamp.
		\protect\subref{subfig:time-encoding}: Two spikes (black vertical lines) from neurons $2$ and $5$.
		\Gls{snn} simulators ---and especially neuromorphic hardware platforms--- often feature high-resolution time encoding.
		However, many \gls{snn} libraries for machine learning frameworks, such as Norse~\citep{pehle2021norse} or others~\citep{manna2023frameworks}, use tensor-like data structures and a time grid that introduces time binning of events (shaded rectangles), often as much as \SI{10}{\percent} of the neurons' membrane time constant, thereby discarding information.
		While sparse data representations can be used to increase coding efficiency, the conversion can be computationally expensive~\citep{spilger2023hxtorchsnn_noeprint}.
		\label{fig:spiketime}
    }
\end{figure}

In particular, for event-based data-sparse training algorithms ---such as EventProp~\citep{wunderlich2021event} or time-to-first-spike approaches~\citep{goeltz2021fast}--- the conversion between event-based and dense time-based representations implies a potential loss of information, due to binning effects on fixed time grids, and performance, see \citet{spilger2023hxtorchsnn_noeprint} for our previous implementation.
While our initial implementation of EventProp for the \gls{bss2} system showed progress towards event-based training~\citep{pehle2023event}, it was still based on a fixed time grid approach using PyTorch's tensor data structures, see \cref{subfig:time-encoding} for a visualization of the event matrix structure.
Recognizing the need for a more streamlined and efficient approach, this work presents a novel event-based software library ---\texttt{jaxsnn}--- for \gls{snn} simulation and for gradient estimation.
It is based on JAX~\citep{jax2022github}, a flexible numerical framework that leverages composable function transformations to provide automatic numerical differentiation for Numpy code (reverse- and forward-mode differentiation), parallelization, vectorization, and support for offloading to hardware accelerators such as \glspl{gpu}.
Our library operates on the neuromorphic native event-based spike representation (see \cref{subfig:spike-encoding}), avoiding the mapping of spike events to a time-discrete grid.
We demonstrate our progress toward event-based modeling using hardware-\acrlong{itl} training with \gls{bss2} on the Yin-Yang dataset~\citep{kriener2021yin}.
The design shift of numerical computation to event-based processing provides a novel way to improve the compatibility with and performance of data-sparse training algorithms, paving the way for more effective neuromorphic computing implementations.

\section{Methods}\label{sec:methods}

\Cref{lst:jaxsnn_idea} presents the basic idea of using JAX to provide a composable modeling interface for machine learning-inspired training on neuromorphic hardware backends.
The apply function defines the forward pass of the \gls{snn} with a given topology.
The computation of spikes is delegated to the neuromorphic software backend, which maps the topology and parameterization to a hardware configuration, triggers the network emulation, captures observables and returns the recorded data.
For hardware \gls{itl} training, this means that we dispatch to the hardware in the ``forward pass'', define an equivalent numerical ``backward'' function, and connect the two functions using \texttt{custom\_jvp}.
This gives us a JAX-transformable function that works on its own;
it can be transformed, composed and chained as needed.
Thus, due to the functional design of JAX, neuromorphic hardware substrates can be naturally integrated and used in \gls{itl} environments.

To test and validate our software framework, we implemented two training algorithms based on previous work~\citep{goeltz2021fast,wunderlich2021event} and applied them to a two-dimensional dataset that is not linearly separable~\citep{kriener2021yin}.
The chosen dataset can be classified using smaller neuromorphic hardware systems.

The \gls{bss2} ASIC~\citep{pehle2022brainscales2_nopreprint_nourl} is a mixed-signal neuromorphic substrate.
It contains 512 neuron compartment circuits (i.e.\ up to 512 single-compartment neurons), each of which implements the AdEx neuron model.
Each compartment circuit which can receive events via 256 synapses.
Events are propagated via digital signals while the post-synaptic neuron dynamics evolve in the analog domain.
The emulation evolves in continuous real time.

\begin{listing}[tb]
	\begin{minted}[fontsize=\footnotesize]{python}
import jax
from mymodel import topology, loss_fn

@jax.custom_vjp
def apply_fn(topology, params, in_spikes):
	return nmhw_backend.bld_and_run(topology, params, in_spikes)

def apply_forward(topology, params, in_spikes):
	# simulate or dispatch experiment to neuromorphic hardware
	rec_data = apply_fn(topology, params, in_spikes)
	return rec_data, (topology, params, in_spikes)

def apply_backward(res, g):
	# topo, params, in_spikes = res
	vjp = calc_grads(res, g)
	return (vjp, None)

apply_fn.defvjp(apply_forward, apply_backward)

# params and in_spikes defined
rec_data = apply_fn(topology, params, in_spikes)
grads = jax.grad(loss_fn)(rec_data, target)
	\end{minted}
	\caption{\label{lst:jaxsnn_idea}%
		jaxsnn functional design principle:
		the forward pass is delegated to the neuromorphic backend and the hardware observables are read out;
		in the backward pass, the gradients are computed, e.g., using the EventProp algorithm~\citep{wunderlich2021event}.
	}
\end{listing}

\subsection{Validation}\label{sec:validation}

\begin{figure}[tb]
	\centering
	\resizebox{.80\linewidth}{!}{%
		\import{.}{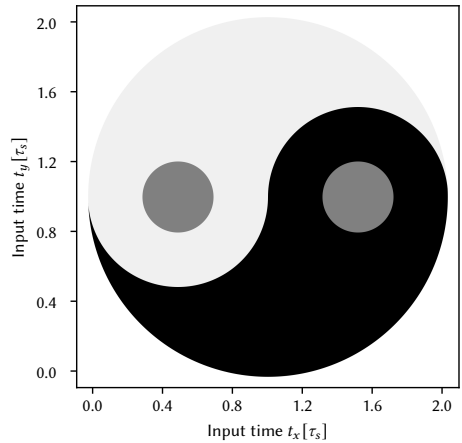}
	}
	\caption{%
		Visualization of the two-dimensional Yin-Yang dataset~\citep{kriener2021yin}.
		The dataset has three different classes (black, white and gray) and is not linearly separable.
		Each two-dimensional data point can be described by the spike times of two input neurons;
		in our experiment, we use mirrored inputs and a bias spike to obtain a five-dimensional input.
		The time axis is defined in relation to the synaptic time constant.
		\label{fig:yinyang-dataset}
	}
\end{figure}

\paragraph{Dataset \& Model}
We trained networks on the Yin-Yang dataset, shown in \cref{fig:yinyang-dataset}.
This dataset was specifically designed for early-stage prototyping and model validation.
As shown in previous publications~\citep{goeltz2021fast,wunderlich2021event,pehle2023event}, good classification results can be achieved with hidden layer sizes of 50 to 150 neurons, making it well suited for, e.g., single-chip \gls{bss2} systems.
The $x$ and $y$ coordinates are mapped to input spike times.
A five-dimensional dataset is obtained by mirroring the resulting spike times with respect to the end of the encoding time interval $t_\text{late}$ and by adding an optional bias spike.
We chose a training set of 5000 samples, and a test set of 3000 samples.

\paragraph{Training Algorithm}
Gradient-based training with spikes from a neuromorphic substrate can benefit from an event-based learning algorithm.
EventProp~\citep{wunderlich2021event} enables gradient-based learning on individual events by computing gradients based on the timing of a spike and the change in membrane voltage at time of the spike.
\citet{pehle2023event} demonstrated learning in a software simulation and on the \gls{bss2} system using this algorithm on our previous top-level machine learning API~\citep{spilger2023hxtorchsnn_noeprint}.
However, this implementation operates on a fixed time grid, a limitation that exists only due to software-technical reasons.
In this work, we provide an event-based implementation of EventProp using our novel library developed on top of JAX.

\section{Results}\label{sec:results}

A mathematical model of hybrid systems ---such as \glspl{snn}--- combines continuous dynamics with transition events.
This model is used to represent complex systems that exhibit both continuous behavior and non-continuous transitions.
A simulation of these systems requires three steps, although the first two are often solved together:

\begin{enumerate}
	\item Find the next event/spike
	\item Solve continuous dynamics to the event time 
	\item Solve non-continuous dynamics/transition 
\end{enumerate}

While the first step can be implemented in software, we support its replacement by emulating the same \gls{snn} on neuromorphic hardware and recording the required observables.

\subsection*{Event-based Simulator}

\Cref{lst:step-algorithm-multiple-neurons} shows the vectorized algorithm for simulating an \gls{snn} with non-delayed recurrent connections and arbitrary topology.
The program jumps from one event to the next event with adjustable steps.
Finding the next event of a \gls{lif} neuron is equivalent to solving for the time $t$ at which $f(x) = V(t) - V_\text{th} = 0$ holds.
The dynamics of the neuron are then integrated up to this point in time.

To support JAX-differentiable and compilable/jittable functions, the memory requirements and thus the number of loop iterations must be computable at compile time.
However, when simulating multiple layers, the number of input spikes for the second layer is already unknown.
We propose to treat input and internal spikes similarly, i.e.\ as transition events with different transition functions.
In each iteration, the simulation steps from one event to the next external (input spike) or internal event (discontinuity of the dynamical system) and $min(t_\text{input}, t_\text{internal})$ is chosen for the time of the next event.
However, specifying the number of events $M$ in advance also has a disadvantage:
it is impossible to simulate a specific time span.
Instead, the simulation will run until a specified number of events, including the input events, have occurred.
There are also two exceptions to handle:
it is not possible to implement a compilable break statement, which is needed if,
first, no next event is found or,
second, the maximum time $t_\text{max}$ is reached.
In this case, a dummy spike will be appended to the list if the network activity is low or if the specified $t_\text{max}$ has been reached.
Since JAX arrays require typed data structures, a dummy spike is represented by an index of $-1$ and a time of \emph{np.inf}, see \cref{subfig:spike-encoding} for the data structure.

To support concurrent simulation, the algorithm is defined in a vectorized fashion:
$V$, $I$ are arrays of size $N$, the number of neurons.
At each step, the index $ix$ of the next spiking neuron is found.
Using an $N \times N$ weight matrix, the algorithm provides support for non-delayed recurrent connections in arbitrary topologies.
This allows multiple layers to be simulated simultaneously.

\begin{algorithm}[tb]
    \caption{\label{lst:step-algorithm-multiple-neurons}%
		Event-based simulation for multiple neurons.
	}
    \begin{algorithmic}[1]
        \For{$i \gets 1$ to $M$}
            \State $t_{\mathrm{internal}} \gets$ find time of next internal event for N neurons
            \State $t_{\mathrm{ix}} \gets \min(t_{\mathrm{internal}})$
            \State $ix \gets \mathrm{argmin} (t_{\mathrm{internal}})$
            \State $t_{\mathrm{external}} \gets$ find time of next input event
            \State $t \gets \min(t_{\mathrm{input}}, t_{\mathrm{ix}})$
            \If{$t > t_{\mathrm{max}}$}
            \State append \textbf{dummy spike} to spike list
            \EndIf
            \State $V, I \gets$ integrate neuron state
            \State append \textbf{spike} to spike list
            \State $V_{ix} \gets V_{\mathrm{reset}}$
            \If{input transition}
                \State $I \gets$ apply input event
            \Else
                \State $I \gets$ apply internal event
            \EndIf
        \EndFor
        \State \Return spike list
    \end{algorithmic}
\end{algorithm}

\begin{figure}[tb]
	\centering
	\subfloat[\label{subfig:step}Step function data flow]{%
		\resizebox{.85\linewidth}{!}{%
			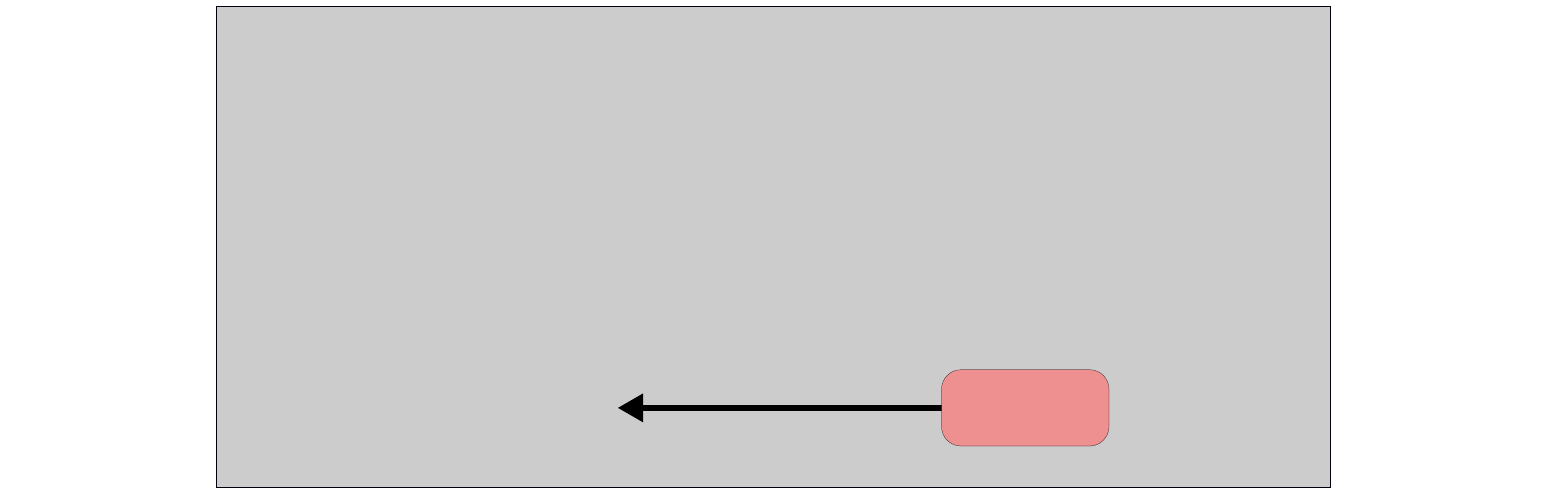%
		}%
	}%
	\\
	\subfloat[\label{subfig:axis-mapping-pytree}Axis mapping]{%
		\resizebox{\linewidth}{!}{%
			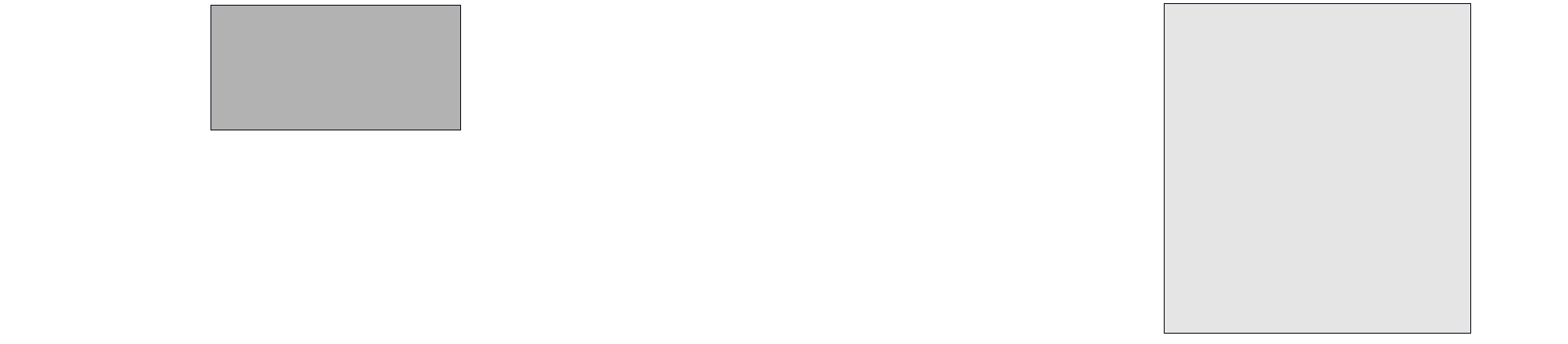%
		}%
	}
	\caption{
		\protect\subref{subfig:step} illustrates the data flow within the \texttt{step} function (cf.\ loop body of \cref{lst:step-algorithm-multiple-neurons}):
		given initial State $S_{t_n}$ and input data the next event time $t_{n+1}$ is determined, the continuous dynamics are progressed to this time $S_{t_{n+1}-}$,
		and the transition/discontinuity is applied, yielding $S_{t_{n+1}}$ and an output spike.
		In \protect\subref{subfig:axis-mapping-pytree},
		each iteration of the \texttt{step} function takes some state $S_t$ and returns an updated state $S_{t+1}$ and some Event $E_{t+1}$.
		When scanning over the \texttt{step} function, the final state $S_\text{last}$ and a \emph{pytree} $\bm{E}$ are returned.
		A new axis is created and mapped down to all leaves of $\bm{E}$.
		The two fields become arrays over their previous type.
	}
\end{figure}

In \cref{subfig:axis-mapping-pytree} we illustrate the data flow in terms of JAX \emph{pytree}\footnote{JAX supports transformations, such as \texttt{scan}, on data structures that adhere to the requirements of \emph{pytree}s.}.
We call the inner part of the for-loop in \cref{lst:step-algorithm-multiple-neurons} \emph{step function} because it takes the \gls{snn} simulation from one state $S_t$ to the next $S_{t+1}$ while generating events $E_{t+1}$.
At the end, it returns the final State $S_{t_\text{last}}$ and a vector of Events $\bm{E}$.
Note that the data access patterns match our neuromorphic/\emph{native} spike data format, previously shown in \cref{subfig:spike-encoding}.

To obtain the next event in the simulation, we use root solving to find the next threshold crossing of the neurons' membrane voltages.
\citet{goeltz2021fast} found analytical solutions to compute the time of the next spike for a \gls{lif} neuron with exponentially decaying current for two special cases $\tau_m = \tau_s$ and $\tau_m = 2 \tau_s$.
Based on these results, we implemented an analytic root solver that is differentiable and provides exact gradients for an \gls{snn} without a \gls{vjp} implementation.
To achieve a JAX-compilable implementation, our implementation relies on JAX primitives (e.g., \texttt{jax.lax.cond} and \texttt{jax.lax.select}) to provide ``control flow'' in terms of branch selection.
We have taken care to provide a \emph{NaN}-safe implementation that allows for JAX vectorization and speculative execution.

Although the developed software can simulate arbitrary topologies and can be extended to multiple neuron types, axonal or synaptic delays in the transmission of spikes between neurons have not been implemented.

\subsection*{Training Algorithms}

We implemented the EventProp algorithm for the \gls{lif} neuron model and synapses with exponentially decaying current\footnote{For example, using the Yin-Yang dataset: \texttt{jaxsnn.event.tasks.yinyang\_event\_prop}}.
The adjoint dynamics of the algorithms are implemented with event-based data structures as a custom \gls{vjp}.
The implementation within our library has been optimized to be compilable to an \gls{xla} representation.
\Gls{xla} is a domain-specific compiler for optimizing and accelerating numerical operations~\citep{tensorflow2015_nourl} originally developed as part of the TensorFlow framework.
It can target various devices such as \glspl{cpu}, \glspl{gpu}, and more specialized hardware backends for numerical acceleration.

In addition, we have implemented an analytical ``\gls{fud}''-based~\citep{goeltz2021fast} training\footnote{\texttt{jaxsnn.event.tasks.yinyang\_analytical}} for unit testing and validation purposes.

\subsection*{Hardware Support}

\begin{figure}[tb]
    \centering
	\resizebox{.9\linewidth}{!}{%
		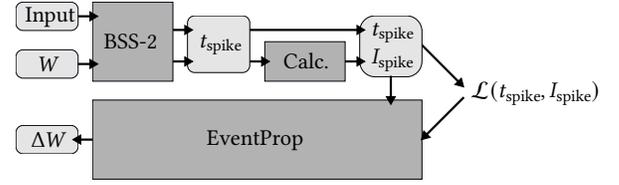%
	}%
    \caption{%
		Hardware \gls{itl} training:
		The forward pass is delegated to the \gls{bss2} system~\citep{pehle2022brainscales2_nopreprint_nourl,mueller2022scalable_noeprint}, which returns a list of spikes.
		Afterwards, an additional step provides information about the synaptic current $I_\text{spike}$ at the time of the spike $t_\text{spike}$, which is numerically calculated.
		Based on a loss function, the EventProp algorithm~\citep{wunderlich2021event} uses these spikes to compute weight updates $\Delta{}W$ for the parameters $W$.
		\label{subfig:itl-training-special}
    }
\end{figure}

Using our custom \gls{vjp} implementation for EventProp, we can replace the numerical root solver/event finder for the \gls{bss2} neuromorphic hardware system.
\Cref{subfig:itl-training-special} outlines this approach in a single step of \gls{itl} training.
By assuming the ideal dynamics (and matching network topology) of the neuromorphic hardware, the backward pass computes exact gradients using only the spikes from the system.

However, the EventProp algorithm we used, relies on spike information and synaptic currents at the same time stamps~\citep{pehle2023event,wunderlich2021event}.
While the currents can be derived from \gls{bss2} system observables, it is not a quantity that can be easily obtained in larger networks.
Instead, we use the spike times from hardware to simulate synaptic currents in software, see $I_\text{spike}$ in \cref{subfig:itl-training-special}.
It would require changes to the hardware substrate or a different training algorithm to eliminate this clunky extra simulation step in the software.
So there are actually two simulated networks learning together.
One that only generates spikes and one that uses these spikes to compute the missing current observable and then performs the backward step.
This is useful not only for evaluating hardware effects, but also for basic validation of the approach.

\newdimen\imageheight

In addition, it may be useful to simulate hardware behavior, for example, to evaluate hardware effects on model behavior in a controlled and gradual manner.
Therefore, it may be beneficial to simulate the hardware backend in software.
In our previous work on the machine learning-inspired modeling \gls{api} \texttt{hxtorch} we introduced a hardware ``mock mode'', which is a simulation of basic hardware effects (e.g., dynamic saturation, parameter variation, etc.).
This way, the event finder can also be replaced by a ``black-box simulation'' incorporating hardware effects.

\subsection*{Experiments}

To validate the new software library and the integration of \gls{bss2} into the training procedure,
we trained the model described in \cref{sec:validation} on the Yin-Yang dataset using EventProp.

For further comparison, we implemented an analytical training algorithm based on \citet{goeltz2021fast},
and an EventProp implementation that operates on a fixed time grid (``\emph{Norse} in \emph{JAX}'').

Our event-based EventProp implementation achieves a test accuracy of \SI{98.2(2)}{\percent} in simulation,
and \SI{94.8(2)}{\percent} on \gls{bss2} using hardware \gls{itl} training.
This is similar to previously published results, see \cref{tab:yy_benchmarks}.

\begin{table}[tb]
	\begin{tabular}{l l l l l   }
		\toprule
		\textbf{Gradient Estimator}           & \textbf{Substrate} & \textbf{Size} & \textbf{Loss} & \textbf{Acc.\ [\%]} \\
		\midrule
		\acrfull{fud}~\citep{goeltz2021fast}     & sim                & 120           & TTFS          & 95.9 \textpm\ 0.7 \\
		\acrfull{fud}~\citep{goeltz2021fast}     & \acrshort{bss2}    & 120           & TTFS          & 95.0 \textpm\ 0.9 \\
		EventProp~\citep{wunderlich2021event} & sim                & 200           & TTFS          & 98.1 \textpm\ 0.2 \\
		\midrule
		EventProp~\citep{pehle2023event}      & sim                & 120           & MOT           & 97.9 \textpm\ 0.6 \\
		EventProp~\citep{pehle2023event}      & \acrshort{bss2}    & 120           & MOT           & 96.1 \textpm\ 0.9 \\
		\midrule
		\emph{Norse} in \emph{JAX}            & sim                & 120           & MOT           & 96.4 \textpm\ 0.2 \\
		\emph{jaxsnn} \acrshort{fud}          & sim                & 120           & TTFS          & 98.1 \textpm\ 0.3 \\
		\emph{jaxsnn} EventProp               & sim                & 120           & TTFS          & 98.2 \textpm\ 0.2 \\
		\emph{jaxsnn} EventProp               & mock               & 100           & TTFS          & 98.0 \textpm\ 0.3 \\
		\emph{jaxsnn} EventProp               & \acrshort{bss2}    & 100           & TTFS          & 94.8 \textpm\ 0.2 \\
		\bottomrule
	\end{tabular}
	\caption{%
		Test accuracies achieved on the Yin-Yang test set after training \glspl{snn}~\citep{goeltz2021fast,wunderlich2021event,pehle2023event} with different gradient estimation methods.
		For loss calculation some models use \gls{ttfs}-encoded spikes from \gls{lif} output neurons, while others use a \gls{mot} membrane voltage from \gls{li} output neurons.
		Most implementations used 120 hidden neurons, \cite{wunderlich2021event} used 200 hidden neurons.
		For runs on \gls{bss2}, or the simulated hardware (``mock mode'') we used 100 hidden neurons.
		\label{tab:yy_benchmarks}
	}
\end{table}

\section{Discussion}\label{sec:discussion}

Our JAX-based~\citep{jax2022github} library~\citep{jaxsnn2023github} is a first step into the realm of event-driven backpropagation-based training libraries for \glspl{snn}.
The modular design of our software serves three purposes:
first, it allows for numerical simulation, gradient computation using Autograd, and gradient-based training of \gls{snn} dynamics without time discretization with support for numerical acceleration using \gls{xla}~\citep{tensorflow2015_nourl}, providing a robust foundation for exploring the intricacies of neural network and training behavior in a controlled, deterministic environment.
Second, the ability to include a ``mock mode'' facilitates behavioral simulations of neuromorphic circuits, providing a testbed for neuromorphic algorithms.
Third, the framework can be seamlessly integrated with neuromorphic hardware backends in the forward pass, which ---in conjunction with hardware \gls{itl}-based training--- provides a bridge between simulation and real-world deployment.

An inherent strength of our design is its modular and composable structure, passing on these inherited qualities from JAX.
This allows the \texttt{jaxsnn} library to be adapted to other neuromorphic hardware platforms, but also positions it for future extensions.
Potential extensions include support for complex neuron dynamics, learning of neuron parameterizations, support for delays, or the incorporation of online plasticity.

A key difference from existing machine learning-inspired \gls{snn} simulation libraries~\citep{fang2023spikingjelly,eshraghian2023training,pehle2021norse,rockpool2019docs} is our explicit focus on event-driven processing.
Our library allows for the direct processing of asynchronous events from neuromorphic platforms.
In cases where the training algorithm does not support asynchronous updates, the processing of densely-encoded spike trains without the need for data conversion can be beneficial in terms of memory consumption and processing speed.

We validated our library, model, and training algorithms on the Yin-Yang dataset.
Using our library, we implemented the analytical training algorithm \acrshort{fud} from \citet{goeltz2021fast}
and an event-based implementation of EventProp~\citep{wunderlich2021event}.
We demonstrated the library for training in an ideal software simulation, on simulated hardware, and on \gls{bss2} single-chip systems.
However, further research is needed,
especially with respect to scaling to larger network sizes
and its impact on training runtime performance and memory requirements,
as well as investigating the effect of high-resolution timing information on result quality,
and ensuring the library's adaptability to a wider range of training algorithms.

Comparing our library with recent advances in cross-platform descriptions of ``\gls{snn} components'' for neuromorphic hardware architectures~\citep{pedersen2023neuromorphic}, certain similarities become apparent.
Our library implements ``units of computation'' that closely resemble the ``\gls{nir} primitives'', and thus has a similar level of abstraction\footnote{This is also true for much older efforts towards a common \gls{snn} experiment description language, e.g.,~\citep{davison2009pynn}. While PyNN describes an \gls{snn} topology as data sources, sinks, and processing units as a graph without time discretization, it does not provide numerical code for simulation, nor does it provide machine learning-friendly data structures.}.
This correspondence positions our library for a straightforward extension to provide gradient estimation functionality for networks described by \gls{nir} primitives.
Such an extension would not only increase the versatility of our library, but could also prove to be a catalyst for community convergence in the cross-platform definition of \glspl{snn}.

\begin{acks}
\label{sec:acknowledgements}
The authors wish to thank all present and former members of the Electronic Vision(s) research group contributing to the \acrlong{bss2} neuromorphic platform.

\noindent%
This work has received funding from
the EC Horizon 2020 Framework Programme
under grant agreement
945539 (HBP SGA3), 
the EC Horizon Europe Framework Programme
under grant agreement
101147319 (EBRAINS 2.0),
and the \foreignlanguage{ngerman}{Deutsche Forschungsgemeinschaft} (DFG, German Research Foundation) under Germany’s Excellence Strategy EXC 2182/1-390900948 (the Heidelberg \mbox{STRUCTURES} Excellence Cluster).
\end{acks}

\section*{Author Contributions}\label{sec:author_contributions}

MA: Investigation, visualization, methodology, software, writing --- original draft, writing --- reviewing \& editing;
CP~\&~EM: Conceptualization, methodology, software, resources, writing — original draft, writing — reviewing \& editing, supervision;
EA~\&~PS: Methodology, software, writing — original draft, writing — reviewing \& editing;
JS: Supervision, funding acquisition, writing — reviewing \& editing.

\bibliographystyle{ACM-Reference-Format}
\bibliography{vision}

\end{document}